\documentclass[runningheads]{llncs}
\usepackage{lineno}  %
\usepackage{dialogue}
\usepackage{tabularx}
\usepackage{amssymb}
\usepackage{amsmath} 
\usepackage{algorithm}
\usepackage{algpseudocode}
\usepackage{float}
\usepackage{mdframed}
\usepackage{booktabs}
\usepackage{xcolor}
\usepackage{threeparttable}
\usepackage[T1]{fontenc}
\usepackage{graphicx}
\usepackage{enumitem}

\newcommand\jj[1]{\textcolor{black}{#1}}

\newcommand{\revision}[1]{\textcolor{black}{#1}}

\begin{document}
\title{Large Language Models as Students Who Think Aloud: Overly Coherent, Verbose, and Confident}
\titlerunning{LLMs as Students: Overly Coherent, Verbose, and Confident}
\author{Conrad Borchers\inst{1} \and
Jill-J\^{e}nn Vie\inst{2} \and
Roger Azevedo\inst{3}}

\authorrunning{Borchers et al.}

\institute{Carnegie Mellon University\\
\email{cborcher@cs.cmu.edu}\\
\and
Soda Team, Inria Saclay
\and
University of Central Florida}
\maketitle 
\begin{abstract}
Large language models (LLMs) are increasingly embedded in AI-based tutoring systems. Can they faithfully model novice reasoning and metacognitive judgments? Existing evaluations emphasize problem-solving accuracy, overlooking the fragmented and imperfect reasoning that characterizes human learning. We evaluate LLMs as novices using 630 think-aloud utterances from multi-step chemistry tutoring problems with problem-solving logs of student hint use, attempts, and problem context. We compare LLM-generated reasoning to human learner utterances under minimal and extended contextual prompting, and assess LLMs' ability to predict step-level learner success. Although GPT-4.1 generates fluent and contextually appropriate continuations, its reasoning is systematically over-coherent, verbose, and less variable than human think-alouds. These effects intensify with a richer problem-solving context during prompting. Learner performance was consistently overestimated. We replicated our findings using the open-source WizardLM-2-7B model. These findings highlight epistemic limitations of simulating learning with LLMs. We attribute these limitations to LLM training data, including expert-like solutions devoid of expressions of affect and working memory constraints during problem solving. Our evaluation framework can guide future design of adaptive systems that more faithfully support novice learning and self-regulation using generative artificial intelligence.
\end{abstract}

\section{Introduction and Related Work}

Large language models (LLMs) are increasingly used as pedagogical agents because they can generate fluent explanations across many domains \cite{team2024learnlm,bubeck2023sparks}. This capacity has motivated their adoption as adaptive tutors. Yet learning sciences research shows that fluent explanations and expert-level knowledge alone do not ensure effective instruction. Experts often misjudge novices by overlooking misconceptions and overestimating prior knowledge, a limitation known as the \emph{expert blind spot} \cite{fisher2016curse,chi1981categorization}.
Understanding learning requires examining learners' reasoning as it unfolds during problem solving to guide effective instruction \cite{koedinger2012knowledge}.
Concurrent think-aloud protocols are especially important for learner modeling as they offer direct, real-time evidence of learners' cognitive and metacognitive processes \cite{ericsson1993protocol}.
In concurrent think-alouds, learners verbalize their thoughts as they solve problems \cite{ericsson1993protocol}. Because these verbalizations avoid the reconstruction biases of retrospective reports, they are especially informative for studying metacognitive monitoring and control. They are therefore widely treated as a benchmark for modeling learner cognition and self-regulated learning \cite{veenman2006metacognition,azevedo2009self}. Think-aloud data also expose clear differences between novice and expert reasoning. Novice verbalizations tend to be fragmented, vague, and focused on surface features, with limited spontaneous monitoring or strategic adjustment \cite{chi1981categorization,pressley2012verbal,hacker2019calibration}. In contrast, expert protocols are more coherent, goal-directed, and explicit about strategy use and integration of information \cite{ericsson2018expertise}.

This raises a key question for LLM-based tutors: when asked to reason step by step as a novice, do models produce the fragmented, uncertain thinking seen in human novices, or do they instead generate polished expert narratives? Because LLMs are pre-trained largely on expert-authored explanations and fine-tuned to provide highly-rated answers, they may have difficulty reflecting novice misconceptions and metacognitive failures \cite{chi1981categorization}. This risk of an expert blind spot could limit their validity as learner models even when told to ``think like a student.'' Despite its importance, there is limited empirical evidence on whether LLM-generated reasoning resembles novice think-alouds or whether contextual prompting improves this alignment.
Progress has been limited by evaluations that prioritize answer accuracy or explanation quality \cite{team2024learnlm,thomas2024tutors} rather than alignment with how novices actually reason and monitor understanding.
Clarifying this issue is critical for guiding the design, evaluation, and use of LLMs in instructional systems that depend on generated reasoning or confidence estimates for adaptation and scaffolding \cite{team2024learnlm,venugopalan2025combining}.

We address this gap by using think-aloud data at the level of problem-solving steps, as a benchmark for whether LLMs resemble novice reasoning.
We compare LLM-generated continuations directly to learner utterances to test whether LLMs follow learners' reasoning.
We also assess metacognitive calibration by evaluating LLMs' ability to predict learners' subsequent problem-solving success and compare these estimates with observed learner outcomes. We ask:

\textbf{RQ1:} How faithfully do LLMs reproduce learner reasoning across different conditions, as measured by similarity to actual student think-alouds?

\textbf{RQ2:} How accurately can LLMs anticipate whether a novice will successfully attempt a problem-solving step, reflecting their metacognitive modeling?

\section{Methods}

We used an open-source dataset of think-aloud chemistry problem solving collected from novice and intermediate students interacting with two intelligent tutoring systems \cite{borchers2024using}.\footnote{Dataset available at \url{https://pslcdatashop.web.cmu.edu/DatasetInfo?datasetId=5371}} The original study included $N=10$ students enrolled in U.S. undergraduate (90\%) and graduate (10\%) programs, recruited from two universities. Participants self-reported moderate prior experience with stoichiometry (mean $=3.4$ on a 5-point scale). Students completed multi-step stoichiometry problems in StoichTutor \cite{mclaren2011polite} while thinking aloud during interactions with the tutoring systems. 

The dataset contains 630 step-level interactions with graded responses \textit{following each} think-aloud utterance (see \cite{borchers2024using}). Our main analyses used GPT-4.1 (Model ID: 1744316542, \revision{run on December 30, 2025)}, which is widely used in the field of AI-based learning technologies \cite{schmucker2024ruffle,borchers2025can}. We replicated the results for RQ1 and RQ2 using the open-source \texttt{WizardLM-2-7B} model \revision{(as a representative, smaller open-source LLM)} via Ollama \cite{xu2023wizardlm}. All analyses and LLM experiments are reproducible using the provided code and can be extended to other models and datasets. \revision{Our open-source code repository also contains details regarding the replication results with WizardLM-2 \cite{repo}. We report results for GPT-4.1 for brevity unless otherwise noted.}
 
\subsection{Modeling Think-Aloud Reasoning (RQ1)}

To address RQ1, we tested how closely LLM-generated continuations resemble human think-aloud reasoning at the level of individual steps. The model generated a single-step think-aloud given the learner's prior activity under two conditions: a simple context with only the immediately preceding learner utterance, and an extended context that also included the problem statement, prior inputs (e.g., ``grams''), which interface element the learner interacted with (e.g., ``first numerator unit''), and tutor feedback. This contrast isolates whether additional context improves the simulation of novice reasoning.

Human think-alouds served as the reference. Human and model utterances were embedded using the \texttt{all-MiniLM-L6-v2} sentence transformer, and similarity was measured with cosine similarity. We compared ground-truth–prediction similarity across conditions using paired randomization tests with 10,000 permutations. To assess coherence, we also compared the similarity between each context and the subsequent human utterance with that between the same context and the model continuation. Higher context–model similarity indicates stronger local coherence. These analyses were conducted separately for simple and extended contexts using identical procedures.

\subsection{Predicting Step-Level Performance (RQ2)}
\label{sec:methods:rq2}

For RQ2, we examined whether LLMs could anticipate learner success in the next step. In two separate experiments, the model produced a probability in [0,1] or a binary correctness judgment. These judgments were included because prior work suggests that LLMs may struggle to generate fine-grained numerical probabilities, as discrete outputs can be more robustly represented in language \cite{zhang20243dg}. Like RQ1, predictions were generated for simple and extended prompts.

Predictions were compared to observed student step correctness from tutor logs. Reliability was assessed by repeating each condition twice to account for stochasticity in generation. Predictive validity was evaluated using point-biserial correlations, and calibration was assessed via calibration bias (a common metacognitive accuracy metric \cite{hacker2013metacognition}), with paired and Welch $t$-tests used to test whether extended context improved performance.

\subsection{Prompt Design}

\revision{When designing prompts, we followed a minimalist approach to (a) observe LLMs natural response to the instruction of thinking aloud and (b) to minimize assumptions. For think-aloud generation, we used a simple instruction prompting the model to act as a novice student and produce the next single think-aloud utterance given the context. For calibration, we used similarly minimal prompts that required the model to output either a probability in $[0,1]$ or a binary correctness judgment for the next step. The prompts define the tasks but do not explicitly encode detailed characteristics of novice reasoning or metacognitive error. This design isolates the model's default behavior and allows us to evaluate how LLMs approximate novice reasoning and performance prediction without additional steering, ensuring that observed differences from human data reflect default LLM tendencies. Accordingly, we used default temperature settings in GPT-4.1. This approach also aligns with our goal of evaluating off-the-shelf behavior of LLMs often deployed in educational settings, e.g. \cite{schmucker2024ruffle}, as opposed to optimizing prompts to induce desired behaviors. Accordingly, our results characterize the extent to which novice-like reasoning and calibration emerge under minimal assumptions, providing a baseline for future work on more structured prompting, fine-tuning, or alignment.}

\section{Results}

\begin{figure}[htpb]
 \centering
 \includegraphics[width=\textwidth]{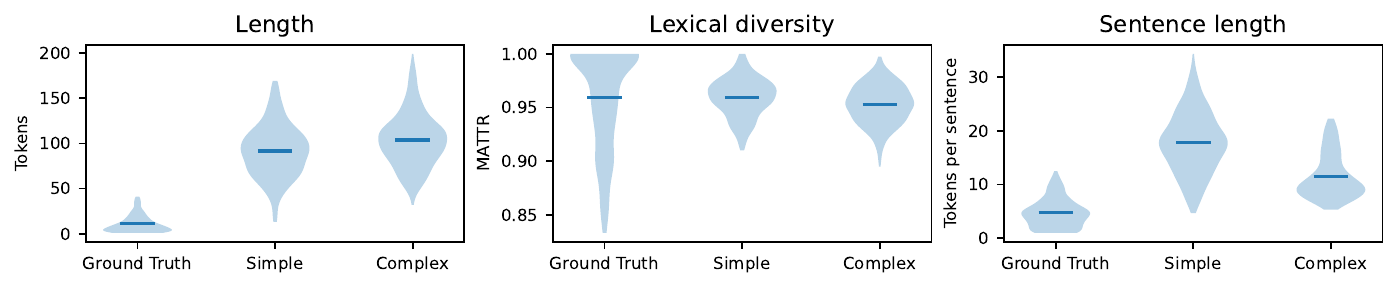}
\caption{\revision{Distributions of linguistic properties for ground-truth learner utterances and model-generated reasoning under simple and extended prompting. Violin plots show per-utterance values, with central markers indicating means. Extreme outliers are removed within each condition for visualization.}}
\label{fig:linguistic_diff}
 \end{figure}

\revision{We start by reporting descriptive analyses of synthetic vs. novice think-aloud.} Figure~\ref{fig:linguistic_diff} compares surface linguistic properties of learner think-alouds with model-generated reasoning under simple and extended prompts. \revision{We analyze response length, mean sentence length, and lexical diversity, measured using the moving-average type--token ratio (MATTR), a length-robust variant of type--token ratio. The figure presents violin plots over per-utterance values; for visualization, extreme outliers are removed within each condition using a median absolute deviation (MAD) filter (values beyond three scaled MADs from the median).} Model-generated explanations are substantially longer and syntactically more elaborate than learner think-alouds. \revision{Lexical diversity is slightly lower for model-generated responses, particularly under extended prompting.} These differences are most pronounced under extended context. Increased contextual input thus encourages fluent, highly coherent explanations that diverge from the exploratory and fragmented reasoning characteristic of novice cognition.

\subsection{RQ1: Fidelity of Think-Aloud Simulation}
\label{sec:res:rq1}

Overall, model-generated think-alouds showed low cosine similarity with human utterances, with higher values indicating greater semantic similarity. Mean similarity was 0.16 for the simple context and 0.19 for the extended context, a small but reliable increase of 0.02 ($p < .001$). Thus, additional context slightly increased similarity. We next examined local coherence by comparing how strongly continuations aligned with their preceding context. In the simple condition, model continuations were much more aligned with prior context than human continuations ($\Delta = -0.22$, $p < .001$). This effect was larger in the extended context condition ($\Delta = -0.44$, $p < .001$). \jj{We replicated those results using the WizardLM-2 model, as documented in our open-source repository \cite{repo}.} These results \revision{provide preliminary evidence} that the model enforces stronger local semantic continuity than learners do, especially when given a richer context. While this produces fluent and consistent continuations, it departs from the variable and exploratory reasoning characteristic of novice problem solving.

To provide intuition, we include a small set of illustrative comparisons between LLM-generated think-alouds and human learner utterances. These patterns were present for both GPT-4.1 and WizardLM-2.
Across both simple and extended context conditions, the LLM frequently generated complete or near-complete expert solutions, even when learner utterances were brief, uncertain, or metacognitive.
For example, a learner's utterance as minimal as \textit{``Two atoms''} elicited a response beginning, \textit{``Alright, let me try to lay this out step by step,''} followed by a formula-based reconstruction of the full problem and a multi-step solution connecting glucose stoichiometry to hydrogen yield. Similarly, \textit{``Let's get the second hint''} prompted a self-contained explanation of the entire solution path \revision{(i.e., \textit{``Alright, let me try to lay this out step by step and see if I have the logic right. I want to find the amount of C6H12O6 (glucose) in moles, and I know that for every 1 mole of glucose, I end up producing 6 moles of H2. So, if I know the total moles of H2 produced, I should divide that by 6 to get the moles of glucose used.'')}.}

\subsection{RQ2: Calibration of Step-Level Performance Predictions}
\label{sec:res:rq2}

We evaluated whether the model produced calibrated estimates of a learner's probability of completing the next step correctly using experimental conditions similar to those in RQ1 (see also Section \ref{sec:methods:rq2}).
\jj{Results are shown in Table~\ref{tab:step_perf} for GPT-4.1 and Table~\ref{tab-wizardlm2_7b}, replicating results for the open-source WizardLM-2 model.}

For probabilistic predictions, GPT-4.1 systematically overestimated learner performance in both contexts (positive calibration bias), with mean errors of $0.325$ in the simple condition and $0.468$ in the extended context condition; both deviations were highly significant (simple: $t=15.56$, $p<.001$; extended: $t=23.92$, $p<.001$). A difference-in-differences analysis showed that extended context increased overestimation by approximately $0.14$ (Welch $t=5.01$, $p<.001$). Discrimination was weak: point-biserial correlations between predicted probabilities and outcomes were not reliable in the simple condition ($r=.031$, $p=.433$) and were small but significant in the extended context condition ($r=.098$, $p=.013$). Despite these calibration and discrimination issues, LLM reliability across repeated runs was high, with correlations across repeated runs ranging from $r=.766$ to $r=.774$. Experiments with WizardLM-2 reproduced this positive calibration bias.

\setlength{\tabcolsep}{6pt}
\begin{table}[htpb]
\centering
\caption{Classification performance for step-level correctness prediction using GPT-4.1. Float predictions are thresholded at 0.50 for Acc./Prec./Rec./F1 and evaluated with ROC AUC on continuous scores; binary predictions do not afford a valid computation of AUC.}
\label{tab:step_perf}
\resizebox{\textwidth}{!}{
\begin{tabular}{llccccccc}
\toprule
Prediction & Context & Acc. & Prec. & Rec. & F1 & AUC & Pred. Accuracy & Calib. Bias \\
\midrule
Float & Simple  & 0.425 & 0.365 & 0.782 & 0.497 & 0.528 & 0.688 & 0.325 \\
Float & Extended & 0.398 & 0.372 & 0.956 & 0.536 & 0.572 & 0.832 & 0.468 \\
\midrule
Binary & Simple  & 0.540 & 0.385 & 0.445 & 0.413 & ---   & 0.421 & 0.057 \\
Binary & Extended & 0.525 & 0.401 & 0.620 & 0.487 & ---   & 0.562 & 0.198 \\
\bottomrule
\end{tabular}
}
\end{table}

\begin{table}[t]
\centering
\caption{Replicated classification performance for step-level correctness prediction \jj{using WizardLM} analogous to Table~\ref{tab:step_perf}.}
\label{tab-wizardlm2_7b}
\resizebox{\textwidth}{!}{
\begin{tabular}{llccccccc}
\toprule
Prediction & Context & Acc. & Prec. & Rec. & F1 & AUC & Pred. Accuracy & Calib. Bias \\
\midrule
Float & Simple  & 0.376 & 0.363 & 0.952 & 0.526 & 0.519 & 0.815 & 0.452 \\
Float & Extended & 0.384 & 0.369 & 0.974 & 0.535 & 0.491 & 0.912 & 0.548 \\
\midrule
Binary & Simple  & 0.392 & 0.361 & 0.873 & 0.511 & ---   & 0.879 & 0.516 \\
Binary & Extended & 0.416 & 0.351 & 0.716 & 0.471 & ---   & 0.741 & 0.378 \\
\bottomrule
\end{tabular}
}
\end{table}

Binary predictions exhibited the same pattern. Paired $t$-tests again indicated overestimation across conditions (simple: $t=2.12$, $p=.034$; extended: $t=7.54$, $p<.001$), with a significant difference-in-differences of $0.141$ (Welch $t=3.75$, $p<.001$). Reliability remained strong ($r=.845$--$.877$), but discrimination was limited, with non-significant correlations in the simple ($r=.038$, $p=.342$) and small correlations in the extended context condition ($r=.089$, $p=.026$).

Overall, across probabilistic and binary outputs, GPT-4.1 produced stable but inflated performance estimates, with extended context amplifying overestimation and yielding only modest improvements in discrimination. Notably, overall classification accuracy was barely above chance. We ran bootstrapping with 10,000 resamples to produce 95\% confidence intervals and found that $AUC$ scores were only marginally better than random guessing (0.50). \jj{This was replicated using the open-source LLM WizardLM-2.} Specifically, simple context predictions achieved $AUC$ = 0.528 [0.480, 0.575] and extended context predictions performed slightly better at $AUC$ = \jj{0.572} [0.525, 0.619] using GPT-4.1.

\section{Discussion}

Our findings indicate that contemporary LLMs, when prompted to generate step-by-step reasoning as novice learners, exhibit systematic limitations as simulators of novice reasoning and metacognitive judgment. Although model-generated think-aloud continuations were fluent and locally appropriate, they were consistently more semantically coherent, verbose, and solution-oriented than authentic learner utterances. Rather than extending learners' fragmented and tentative reasoning states, models tended to impose smooth, globally consistent narratives that diverge from the exploratory nature of novice problem solving. This pattern parallels the expert blind spot documented in cognitive psychology \cite{fisher2016curse}. Like human experts, LLMs reconstruct polished solution paths that omit hesitation, misconception, and partial understanding, potentially because their training data distribution represents expert-authored solutions to problems (e.g., textbook proofs). Richer contextual prompting amplified this tendency, increasing over-coherence and verbosity by encouraging LLMs to resolve ambiguities that human novices typically leave unresolved. These results challenge the assumption that additional context necessarily produces more pedagogically realistic model behavior \cite{venugopalan2025combining}.

A similar misalignment emerged in metacognitive modeling (RQ2). When predicting step-level learner success, LLMs produced poorly calibrated estimates that systematically overestimated performance, even after re-prompting. This overconfidence intensified with extended context and risks driving instructional decisions that are overly advanced or prematurely faded, thereby undermining effective scaffolding and adaptive support \cite{koedinger2007exploring}.

Several structural factors help explain why LLM-generated think-alouds diverge from human novices. First, LLMs lack the cognitive constraints that shape human verbalization during problem solving. Human think-alouds are bounded by working-memory limitations and attentional bottlenecks, which naturally produce pauses, incomplete utterances, and breakdowns in articulation under difficulty \cite{ericsson1993protocol}. In contrast, LLMs generate text through constrained sequence completion that favors coherence learned from expert solutions, so over-coherence reflects the absence of constraints that produce fragmented novice reasoning. Second, LLMs do not possess experienced affective or motivational states. Novice think-alouds frequently include expressions of frustration and self-doubt that influence strategy selection and monitoring accuracy. In contrast, emotional language in LLM outputs reflects learned linguistic patterns rather than experienced regulation \cite{demszky2023using}. This limits LLMs' ability to reproduce authentic self-regulated learning dynamics. Third, LLMs do not undergo learning or conceptual change during problem solving in the human sense. Whereas human novices exhibit gradual knowledge construction and occasional insight-driven restructuring, learning trajectories in LLM outputs must be simulated via prompting \cite{carey1985conceptual}.

\subsection{Limitations and Future Work}

This study has three primary limitations. First, our analyses are based on a single dataset of chemistry tutoring interactions. While chemistry provides a canonical setting for stepwise problem solving \cite{borchers2024using}, the extent to which over-coherence and miscalibration generalize to other domains and instructional contexts remains an open question. Second, we evaluated general-purpose LLMs operating under standard prompting, without fine-tuning or alignment with pedagogical principles \cite{team2024learnlm}. Accordingly, our results characterize the behavior of contemporary, off-the-shelf models as they are currently deployed in educational systems \cite{schmucker2024ruffle,venugopalan2025combining}, rather than the theoretical limits of LLM-based learner modeling. Third, our evaluation focused on step-level predictions and continuations in isolation. \revision{Considering more complex prompting techniques, for instance, longer think-aloud utterance sequences, might} enable LLMs to adapt more effectively.

Our findings indicate that addressing the over-coherence and miscalibration of LLM-generated think-alouds requires explicit constraints on how novice cognition is modeled. Faithful simulation depends on grounding generation in empirically documented novice knowledge gaps and misconceptions \cite{chi1981categorization,ericsson1993protocol}. This includes incorporating realistic cognitive-load limits that yield fragmented and imprecise verbalizations under difficulty \cite{sweller2023development}, and reproducing novice-typical metacognitive miscalibration such as overconfidence and weak error detection \cite{koriat2005illusions}. Achieving this fidelity will likely require fine-tuning on authentic think-aloud data and validation against human think-aloud protocols, as opposed to instruction-following regimes that primarily optimize for adherence to instructional principles (e.g., building on the learner's prior knowledge before giving instruction) from an expert perspective, as seen in current ``pedagogical LLMs'' \cite{team2024learnlm}.

\section{Conclusion}

This study contributes a novel evaluation framework for assessing LLMs as simulators of novice reasoning by grounding analysis in authentic think-aloud data. Using step-level comparisons between model-generated and human learner utterances, we show that although LLMs produce fluent and contextually appropriate reasoning, they do not reflect the fragmented, uncertain, and error-prone nature of novice thinking. Instead, their outputs are systematically over-coherent, verbose, and solution-oriented, and their step-level predictions of learner success are poorly calibrated and consistently overconfident. These patterns persist across prompting conditions and are replicated in both proprietary and open-source models.

Theoretically, our findings highlight a fundamental mismatch between the statistical regularities learned by LLMs and the cognitive processes that characterize human learning. In particular, they suggest that novice reasoning cannot be approximated by simply prompting models to ``think like a student,'' as LLMs default to expert-like representations that omit key features of learning, including uncertainty, partial understanding, and metacognitive error. This extends prior work on the expert blind spot by demonstrating an analogous limitation in generative models and reinforce the importance of evaluating AI systems not only on accuracy or fluency, but on their alignment with human cognitive processes.

Practically, these results have direct implications for the design of AI-based tutoring systems. Overly coherent reasoning and overestimation of learner performance may lead to inappropriate instructional decisions, such as prematurely advancing content or reducing necessary scaffolding. Our findings therefore caution against using off-the-shelf LLMs as learner models without explicit constraints or empirical validation. More broadly, they motivate the development of approaches that incorporate realistic cognitive limitations, learner variability, and metacognitive dynamics, for example through fine-tuning on think-aloud data or hybrid systems that combine generative models with structured learner modeling techniques.

Finally, we emphasize the need for richer empirical benchmarks to support this line of work. We encourage the release of additional open think-aloud datasets and standardized evaluation protocols to enable more robust, generalizable assessments of how well AI systems capture the processes of human learning, metacognition, and self-regulation.

\bibliographystyle{splncs04}
\bibliography{main}

\end{document}